\documentclass[10pt,twocolumn,letterpaper]{article}

\usepackage[pagebackref=true,breaklinks=true,letterpaper=true,colorlinks,bookmarks=false]{hyperref}
\usepackage{cvpr}
\usepackage{times}
\usepackage{epsfig}
\usepackage{graphicx}
\usepackage{amsmath}
\usepackage{amssymb}
\usepackage{multirow}
\usepackage{makecell}
\usepackage{mmstyles}
\usepackage{bbding}
\usepackage{subcaption}

\newenvironment{packed_itemize}{
    \vspace{-0.15cm}\begin{itemize}
        \setlength{\itemsep}{1pt}
        \setlength{\parskip}{0pt}
        \setlength{\parsep}{0pt}
    }{\end{itemize}}

\cvprfinalcopy 

\begin{document}

\title{Video Object Segmentation with Joint Re-identification and \\ Attention-Aware Mask Propagation}

	\author{Xiaoxiao Li \hspace{9pt} Chen Change Loy \\
		\small{Department of Information Engineering, The Chinese University of Hong Kong}\\
		{\tt\small \{lx015,ccloy\}@ie.cuhk.edu.hk}
	}

\maketitle

\begin{abstract}
The problem of video object segmentation can become extremely challenging when multiple instances co-exist. While each instance may exhibit large scale and pose variations, the problem is compounded when instances occlude each other causing failures in tracking. 
%
In this study, we formulate a deep recurrent network that is capable of segmenting and tracking objects in video simultaneously by their temporal continuity, yet able to re-identify them when they re-appear after a prolonged occlusion. We combine both temporal propagation and re-identification functionalities into a single framework that can be trained end-to-end. In particular, we present a re-identification module with template expansion to retrieve missing objects despite their large appearance changes. In addition, we contribute a new attention-based recurrent mask propagation approach that is robust to distractors not belonging to the target segment. Our approach achieves a new state-of-the-art global mean (Region Jaccard and Boundary F measure) of 68.2 on the challenging DAVIS 2017 benchmark~\cite{pont2017the} (test-dev set), outperforming the winning solution~\cite{li2017video} which achieves a global mean of 66.1 on the same partition\footnote{Li \etal~\cite{li2017video} also reported their performances on \textit{test-challenge} set, however, the \textit{test-challenge} entry is now closed. The $\mathcal{G}$-mean on \textit{test-challenge} set is usually  $2\sim3$ points higher than $\mathcal{G}$-mean on \textit{test-dev}.}.
\end{abstract}

\section{Introduction}
\label{sec:intro}

\begin{figure}[t]
\begin{center}
\includegraphics[width=0.48\textwidth]{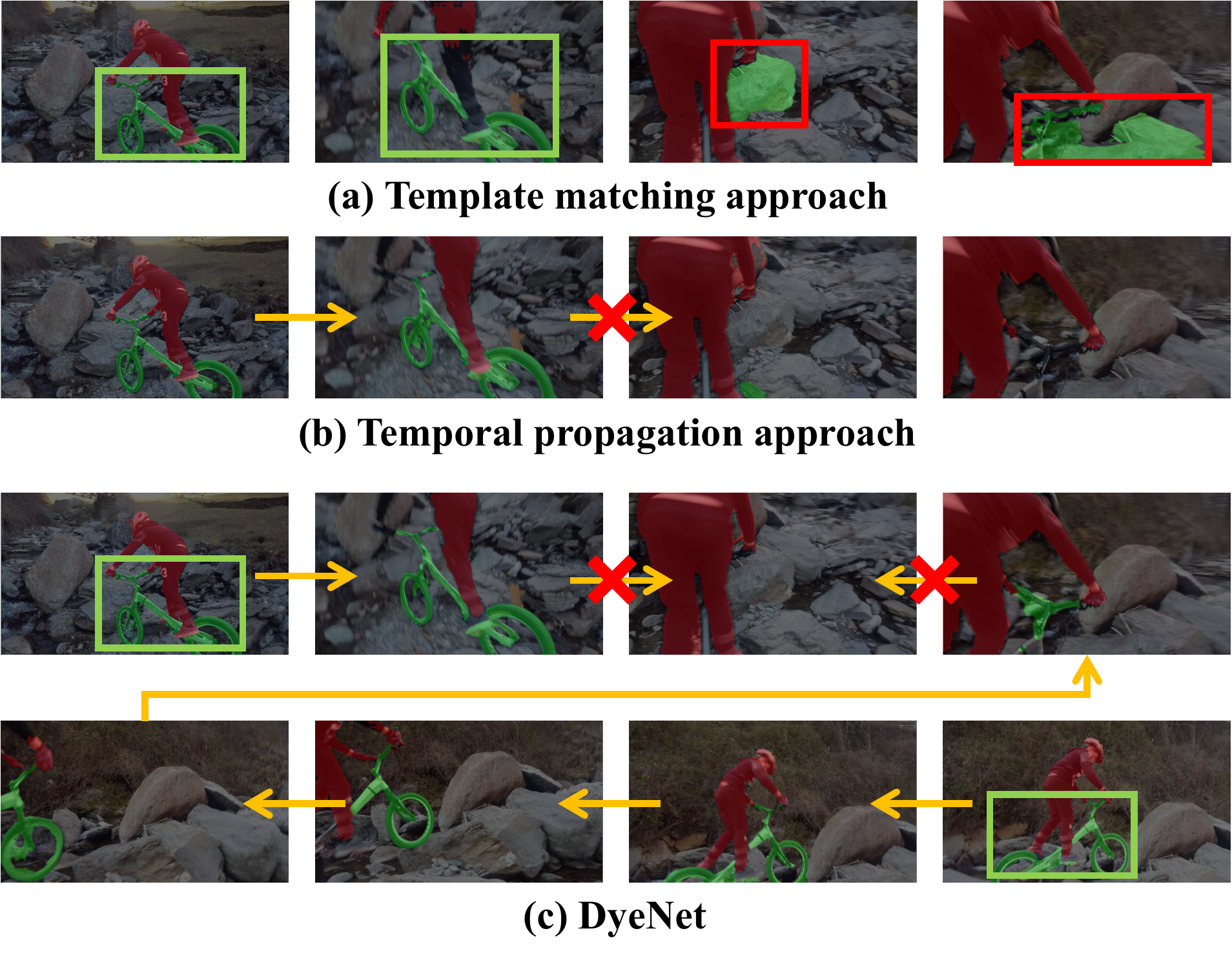}
\vskip -0.4cm
\caption{\small{In this example, we focus on the bicycle. (a) shows the result of template matching approach which is affected by large scale and pose variations. As shown in (b), the temporal propagation approach is incapable of handling occlusion. The proposed DyeNet joints them into a unified framework, first retrieves high confidence starting points and then propagates their masks bidirectionally to address those issues. The result of DyeNet is visualized in (c).
\textbf{Best viewed in color.}}}
\label{fig:overview}
\end{center}
\vspace{-20pt}
\end{figure}

Video object segmentation aims at segmenting foreground instance object(s) from the background region in a video sequence. Typically, ground-truth masks are assumed to be given in the first frame. The goal is to begin with these masks and track them in the remaining sequence. This paradigm is sometimes known as semi-supervised video object segmentation in the literature~\cite{marki2016bilateral,caelles2017one,perazzi2016benchmark}.
A notable and challenging benchmark for this task is 2017 DAVIS Challenge
~\cite{pont2017the}. An example of a sequence is shown in Fig.~\ref{fig:overview}.
The DAVIS dataset presents real-world challenges that need to be solved from two key aspects. First, there are multiple instances in a video. It is very likely that they will occlude each other causing partial or even full obstruction of a target instance. Second, instances typically experience substantial variations in both scale and pose across frames. 

To address the occlusion problem, notable studies such as~\cite{caelles2017one,yoon2017pixel} adapt generic semantic segmentation deep model to the task of specific object segmentation.
These methods follow a notion reminiscent of the template matching based methods that are widely used in visual tracking task~\cite{bolme2010visual,valmadre2017end}. Often, a fixed set of templates such as the masks of target objects in the first frame are used for matching targets. 
%
%
This paradigm fails in some challenging cases in DAVIS (see Fig.~\ref{fig:overview}(a)), as using a fixed set of templates cannot sufficiently cover large scale and pose variations.
%
%
%
To mitigate the variations in both scale and pose across frames, existing studies~\cite{tsai2016video,xiao2016track,jampani2017video,perazzi2017learning,voigtlaender2017online,khoreva2017lucid} exploit temporal information to maintain continuity of individual segmented regions across frames.
%
%
%
On unconstrained videos with severe occlusions, such as that shown in Fig.~\ref{fig:overview}(b), approaches based on temporal continuity are prone to errors since there is no mechanism to re-identify a target when it
reappears after missing in a few video frames. In addition, these approaches may fail to track instances in the presence of distractors such as cluttered backgrounds or segments from other objects during temporal propagation.

Solving video object segmentation with multiple instances requires template matching for coping with occlusion and temporal propagation for ensuring temporal continuity. 
In this study, we bring both approaches into a single unified network.
Our network hinges on two main modules, namely a re-identification (Re-ID) module and a recurrent mask propagation (Re-MP) module.
The Re-ID module helps to establish confident starting points in non-successive frames and retrieve missing segments caused by occlusions.
Based on the segments provided by the Re-ID module, the Re-MP module propagates their masks bidirectionally by a recurrent neural network to the entire video.
The process of conducting Re-ID followed by Re-MP may be imagined as dyeing a fabric with multiple color dots (\ie,~choosing starting points with re-identification) and the color disperses from these dots (\ie,~propagation).
Drawing from this analogy, we name our network as \textit{DyeNet}.

There are a few methods~\cite{nguyen2017instance,li2017video} that improve video object segmentation through both temporal propagation and re-identification. Our approach differs by offering a unified network that allows both tasks to be optimized in an end-to-end network.
In addition, unlike existing studies, the Re-ID and Re-MP steps are conducted in an iterative manner. This allows us to identify confidently predicted mask in each iteration and expand the template set. With a dynamic expansion of template set, our Re-ID module can better retrieve missing objects that reappear with different poses and scales. 
In addition, the Re-MP module is specially designed with attention mechanism to disregard distractors such as background objects or segments from other objects during mask propagation. 
As shown in Fig.~\ref{fig:overview}(c), DyeNet is capable of segmenting multiple instances across a video with high accuracy through Re-ID and Re-MP. We provide a more detailed discussion against \cite{nguyen2017instance,li2017video} in the related work section.

Our \textbf{contributions} are summarized as follows. 
(1) We propose a novel approach that joints template matching and temporal propagation into a unified deep neural network for addressing video object segmentation with multiple instances. The network can be trained end-to-end. It does not require online training (\ie, fine-tune using the masks of the first frame) to do well but can achieve better results with online training.
(2) We present an effective template expansion approach to better retrieve missing targets that reappear with different poses and scales.
(3) We present a new attention-based recurrent mask propagation module that is more resilient to distractors.

We use the challenging DAVIS 2017 dataset~\cite{pont2017the} as our key benchmark. 
The winner of this challenge~\cite{li2017video} achieves a global mean (Region Jaccard and Boundary F measure) of 66.1 on the test-dev partition.
%
Our method obtains a global mean of 68.2 on this partition. Without online training, DyeNet can still achieve a competitive $\mathcal{G}$-mean of $62.5$ while the speed is an order of magnitude faster.
Our method also achieves state-of-the-art results on DAVIS 2016~\cite{perazzi2016benchmark}, SegTrack$_{\rm v2}$~\cite{FliICCV2013} and YouTubeObjects~\cite{PrestLCSF12} datasets.

\section{Related Work}
\label{sec:related_work}

\noindent
\textbf{Image segmentation.}
%
%
The goal of semi-supervised video object segmentation is different to semantic image segmentation~\cite{chen2014semantic,crfasrnn_iccv2015,liu2017deep,li2017not,zhao2017pspnet} and instance segmentation~\cite{hariharan2014simultaneous,hariharan2015hypercolumns,li2016fully,he2017mask} that perform pixel-wise class labeling. In video object segmentation, the class type is always assumed to be undefined. Thus, the challenge lies in performing accurate object-agnostic mask propagation. 
Our network leverages semantic image segmentation task to learn generic representation that encompasses semantic level information. 
%
%
%
The representation learned is strong, allowing our model to be applied in a dataset-agnostic manner, \ie, it is not trained with any first frame annotation of each video in the target dataset as training/tuning set, but it can also be optionally fine-tuned and adapted into the targeted video domain as practiced in~\cite{khoreva2017lucid} to obtain better results.
We will examine both possibilities in the experimental section.

\noindent
\textbf{Visual tracking.}
While semi-supervised video object segmentation can be seen as a pixel-level tracking task, video object segmentation differs in its more challenging nature in terms of object scale variation across video frames and inter-object scale differences. In addition, the pose of objects is relatively stable in the tracking datasets, and there are few prolonged occlusions. Importantly, the problem differs in that conventional tracking tasks only need bounding box level tracking results, and concern about causality (\ie, tracker does not use any future frames for estimation).
In contrast, semi-supervised video object segmentation expects precise pixel-level tracking results, and typically does not assume causality.


\noindent
\textbf{Video object segmentation.}
%
%
Prior to the prevalence of deep learning, most approaches to semantic video segmentation are graph based~\cite{grundmann2010efficient,lee2011key,xu2012streaming,papazoglou2013fast}. 
Contemporary methods are mostly based on deep learning. 
%
A useful technique reminiscent of template matching is commonly applied.
In particular, templates are typically formed by the ground-truth masks in the first frame.
For instance, Caelles~\etal~\cite{caelles2017one} adapt a generic semantic image segmentation network to the templates for each testing video individually.
Yoon~\etal~\cite{yoon2017pixel} distinguish the foreground objects based on the pixel-level similarity between candidates and templates, which is measured by a matching deep network.
Another useful technique is to exploit temporal continuity for establishing spatiotemporal correlation. 
Tsai \etal~\cite{tsai2016video} estimate object segmentation and optical flow synergistically using an iterative scheme.
Jampani \etal~\cite{jampani2017video} propagate structured information through a video sequence by a bilateral network that performs learnable bilateral filtering operations cross video frames.
Perazzi \etal~\cite{perazzi2017learning} and Jang \etal~\cite{voigtlaender2017online} estimate the segmentation mask of the current frame by using the mask from the previous frame as a guidance.
%

\noindent
\textbf{Differences against existing methods that combine template matching and temporal continuity.}
There are a few studies that combine the merits of the two aforementioned techniques.
Khoreva \etal~\cite{khoreva2017lucid} show that a training set closer to the target domain is more effective. They improve~\cite{caelles2017one} by synthesizing more training data from the first frame of testing videos
%
and employ mask propagation during the inference.
Instance Re-Identification Flow (IRIF)~\cite{nguyen2017instance} divides foreground objects into human and non-human object instances, and then apply person re-identification network~\cite{xiao2017joint} to retrieve missing human during mask propagation and blend them into the final prediction. For non-human object instances, IRIF degenerates to a conventional mask propagation method.
Our method differs to these studies in that we do not synthesize training data from the first frames and do not explicitly divide foreground objects into human and non-human object instances.

Li \etal~\cite{li2017video} adapt person re-identification approach~\cite{xiao2017joint} to a generic object re-identification model and employ a two-stream mask propagation model~\cite{perazzi2017learning}.
Their method (VS-ReID) achieved the highest performance in the 2017 DAVIS Challenge~\cite{li2017video}, however, its shortcomings are also obvious:
(1) Unlike DyeNet that adopts template expansion, VS-ReID only uses the masks of target objects in the first frame as templates. It is thus more susceptible to pose variations.
(2) Their method is much slower compared to ours due to its redundant feature extraction steps and less efficient inference method. Specifically, the inference of VS-ReID takes $\sim$3 seconds per frame on DAVIS dataset. The speed is 7 times slower than DyeNet.
(3) VS-ReID does not have any attention mechanism in its mask propagation. Its robustness to distractors and background clutters is thus inferior to DyeNet. 
(4) VS-ReID cannot perform end-to-end training. By contrast, DyeNet performs joint learning of re-identification and temporal propagation.
%
%
%

\section{Methodology}
\label{sec:approach}

\begin{figure*}[t]
	\centering
	\includegraphics[width=0.95\textwidth]{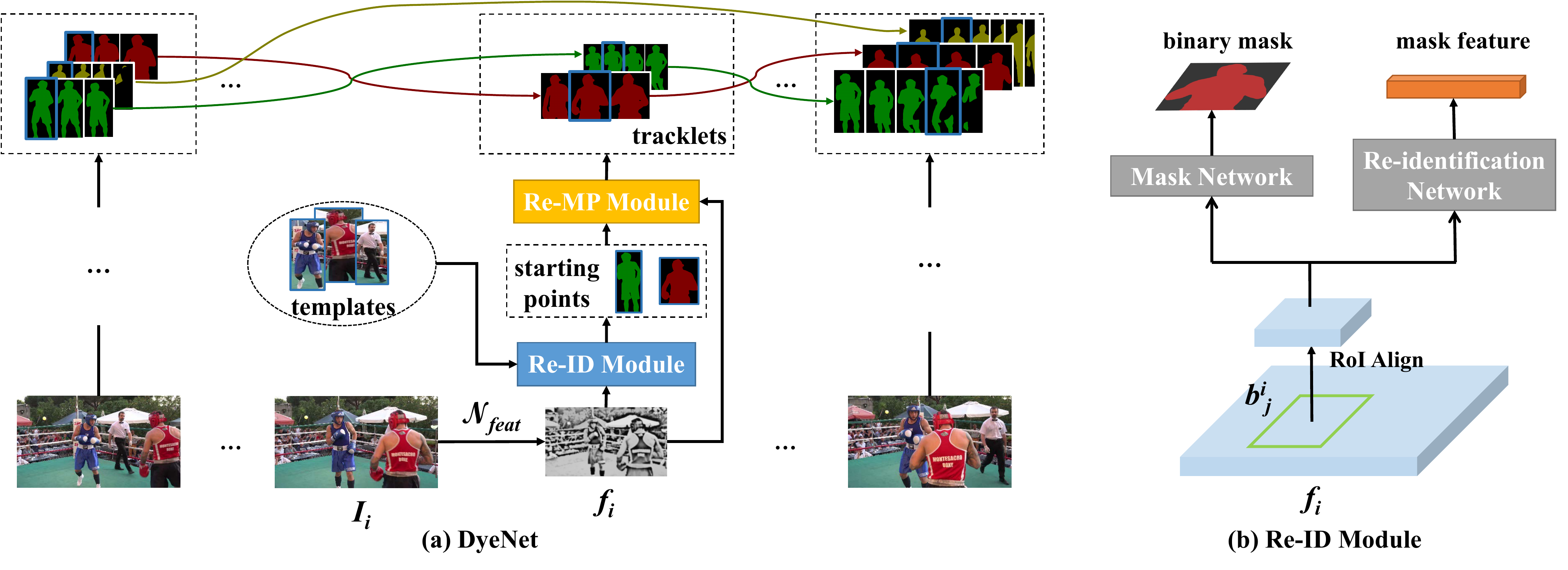}
	\vskip -0.4cm
	\caption{\small{(a) The pipeline of DyeNet. The network hinges on two main modules, namely a re-identification (Re-ID) module and a recurrent mask propagation (Re-MP) module. (b) The network architecture of the re-identification (Re-ID) module. \textbf{Best viewed in color.}}}
	\label{fig:model_architecture}
	\vspace{-10pt}
\end{figure*}

We provide an overview of the proposed approach.
%
Figure~\ref{fig:model_architecture}(a) depicts the architecture of DyeNet. It consists of two modules, namely the re-identification (Re-ID) module and the recurrent mask propagation (Re-MP) module.
The network first performs feature extraction, which will be detailed next.

\vspace{0.1cm}
\noindent
\textbf{Feature extraction}.
Given a video sequence with $N$ frames $\{I_1,...,I_N\}$, for each frame $I_i$, we first extract a feature $f_i$ by a convolutional feature network $\mathcal{N}_{feat}$, \ie,~$f_i = \mathcal{N}_{feat}(I_i)$.
Both Re-ID and Re-MP modules employ the same set of features in order to save computation in feature extraction.
Considering model capacity and speed, we use ResNet-101~\cite{he2016deep} as the backbone of $\mathcal{N}_{feat}$.
More specifically, ResNet-101 consists of five blocks named as `conv1', `conv2\_x' to `conv5\_x'.
We employ `conv1' to `conv4\_x' as our feature network.
To increase the resolution of features, we decrease the convolutional strides in `conv4\_x' block and replace convolutions in `conv4\_x' by dilated convolutions similar to~\cite{chen2014semantic}.
Consequently, the resolution of feature maps is $1/8$ of the input frame.

\vspace{0.1cm}
\noindent
\textbf{Iterative inference with template expansion}.
After feature extraction, DyeNet runs Re-ID and Re-MP in an iterative manner to obtain segmentation masks of all instances across the whole video sequence.
We assume the availability of masks given in the first frame and use them as templates. This is the standard protocol of
the benchmarks considered in Sec.~\ref{sec:experiments}.


In the first iteration, the Re-ID module generates a set of masks from object proposals and compares them with templates. Masks with a high similarity to templates are chosen as the starting points for Re-MP.
Subsequently, Re-MP propagates each selected mask (\ie, starting point) bidirectionally, and generates a sequence of segmentation masks, which we call tracklet. 
After Re-MP, we can additionally consider post-processing steps to link the tracklets.
In subsequent iterations, DyeNet chooses confidently predicted masks to expand the template set and reapplies Re-ID and Re-MP.
Template expansion avoids heavy reliance on the masks provided by the first frame, which may not capture sufficient pose variations of targets.
%

Note that we do not expect to retrieve all the masks of target objects in a given sequence.
In the first iteration, it is sufficient to obtain several high-quality starting points for the mask propagation step.
After each iteration of DyeNet, we select predictions with high confidence to augment the template set. 
In practice, the first iteration can retrieve nearly $25\%$ masks as starting points on DAVIS 2017 dataset.
After three iterations, this rate will increase to $33\%$. In this work, DyeNet stops the iterative process when no more high-confident masks can be found by the Re-ID module.
Next, we present the Re-ID and Re-MP modules.

\begin{figure}[b]
	\centering
	\includegraphics[width=0.48\textwidth]{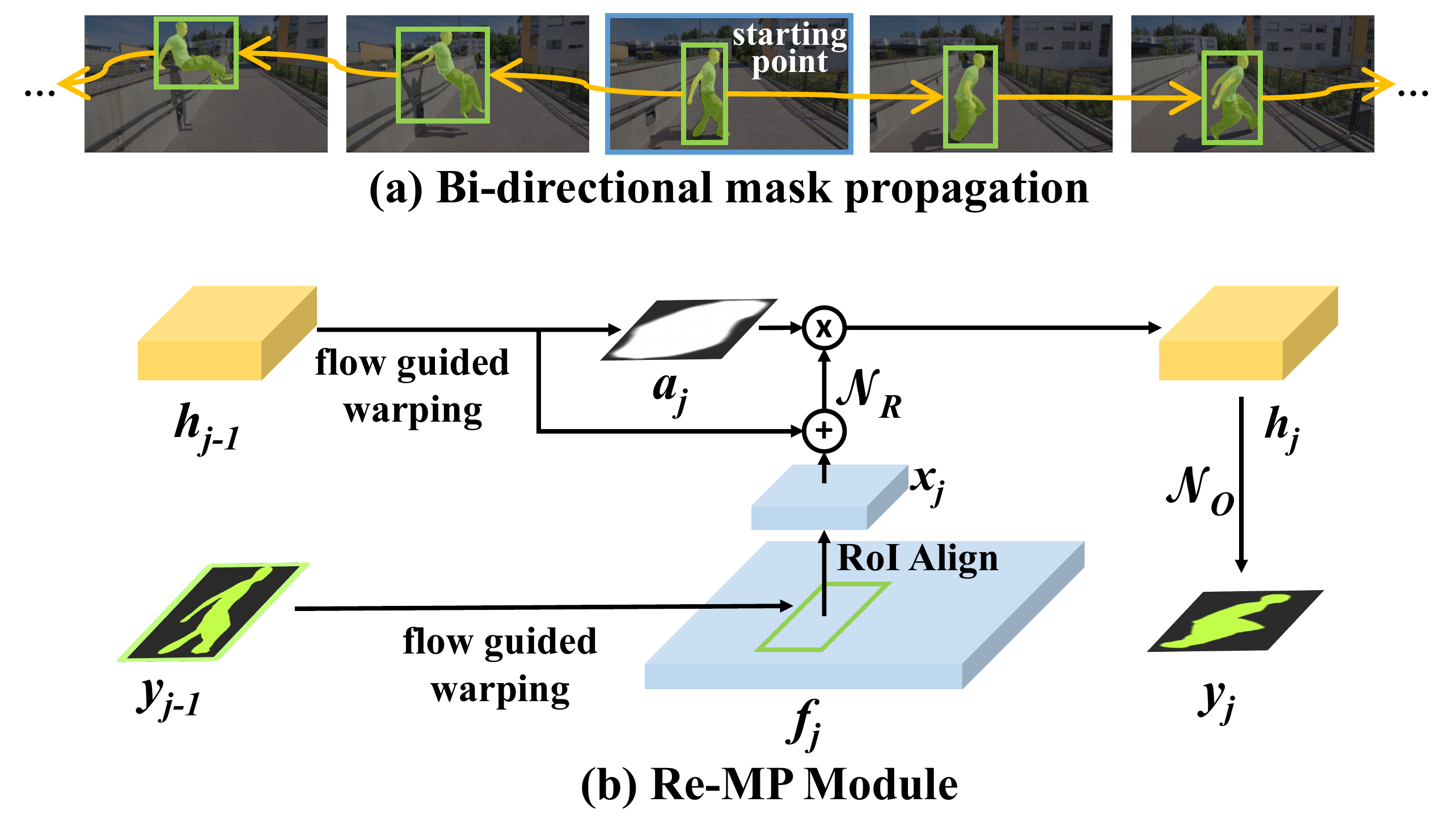}
	\vskip -0.2cm
	\caption{\small{(a) Illustration of bi-direction mask propagation. (b) The network architecture of the recurrent mask propagation (Re-MP) module. \textbf{Best viewed in color.}}}
	\label{fig:mp}
	\vspace{-15pt}
\end{figure}

\subsection{Re-identification}
\label{sec:re_identification}

We introduce the Re-ID module to search for targets in the video sequences. The module has several unique features that allow it to retrieve a missing object that reappears in different scales and poses.
First, as discussed previously, we expand the template set in every iteration we apply Re-ID and Re-MP. Template expansion enriches the template set for more robust matching.
Second, we employ the object proposal method to estimate the location of target objects.
Since these proposals are generated based on anchors of various sizes, which cover objects of various scales, the Re-ID module can handle large scale variations.

Figure~\ref{fig:model_architecture}(b) illustrates the Re-ID module.
For the $i$-th frame, besides the feature $f_i$, the Re-ID module also requires the object proposals  $\{b^i_1,...,b^i_M\}$ as input where $M$ indicates the number of proposal bounding boxes on this frame.
We employ a Region Proposal Network (RPN)~\cite{ren2015faster} to propose candidate object bounding boxes on every frame.
For convenience, our RPN is trained separately from DyeNet, but their backbone networks are shareable.
For each candidate bounding box $b^i_j$, we first extract its feature from $f_i$, and resize the feature into a fixed size $m \times m$ (\eg, 28$\times$28) by RoIAlign~\cite{he2017mask}, which is an improved form of RoIPool that removes harsh quantization.
The extracted features are fed into two shallow sub-networks.
The first sub-network is a mask network that predicts a $m \times m$ binary mask that represents the segmentation mask of the main instance in candidate bounding box $b^i_j$.
The second sub-network is a re-identification network that projects the extracted features into an L2-normalized 256-dimensional subspace to obtain the mask features.
The templates are also projected onto the same subspace for feature extraction.

By computing the cosine similarities between the mask and template features, we can measure the similarity between candidate bounding boxes and templates.
If a candidate bounding box is sufficiently similar to any template, that is, the cosine similarity is larger than a threshold $\rho_{reid}$, we will keep its mask as a starting point for mask propagation.
In practice, we set $\rho_{reid}$ with a high value to establish high-quality starting points for our next step.

We employ `conv5\_x' block of ResNet-101 as the backbone of the sub-networks.
However, some modifications are necessary to adapt them to the respective tasks.
In particular, we decrease the convolutional strides in the mask network to capture more details of prediction.
For the re-identification network, we keep the original strides and append a global average pooling layer and a fully connected layer to project the features into the target subspace.

\subsection{Recurrent Mask Propagation}
\label{sec:recurrent_mp}

As shown in Fig.~\ref{fig:mp}(a), we bi-directionally extend the retrieved masks (\ie, starting points) to form tracklets by using the Re-MP module.
By incorporating short-term memory, the module is capable of handling large pose variations, which complements the re-identification module.
We formulate the Re-MP module as a Recurrent Neural Network (RNN).
Figure~\ref{fig:mp}(b) illustrates the mask propagation process between adjacent frames.
For brevity, we only describe the forward propagation. A backward propagation can be conducted with the same approach.

Suppose $\hat{y}$ is a retrieved segmentation mask for instance $k$ in the $i$-th frame, and we have propagated $\hat{y}$ from $i$-th frame to $(j-1)$-th frame, $\{y_{i+1},y_{i+2}, ..., y_{j-1}\}$ is the sequence of binary masks that we obtain.
We now aim to predict $y_{j}$, \ie, the mask for instance $k$ in the $j$-th frame.
In a RNN framework, the prediction of $y_{j}$ can be solved as 
\begin{eqnarray}
\label{eqn:hidden_rnn}
{h_j} & = & \mathcal{N}_{R}(h_{(j-1) \to j}, x_j), \\ 
\label{eqn:output_rnn}
y_{j} & = & \mathcal{N}_{O}(h_j),
\end{eqnarray}
where $\mathcal{N}_{R}$ and $\mathcal{N}_{O}$ are the recurrent function and output function, respectively.

We first explain Eq.~\eqref{eqn:hidden_rnn}. 
We begin with estimating the location, \ie,~the bounding box, of instance $k$ in the $j$-th frame from $y_{j-1}$ by flow guided warping.
More specifically, we use FlowNet 2.0~\cite{ilg2017flownet} to extract the optical flow $F_{(j-1) \to j}$ between $(j-1)$-th and $j$-th frames.
The binary mask $y_{j-1}$ is warped to $y_{(j-1) \to j}$ according to $F_{(j-1) \to j}$ by a bilinear warping function.
After that, we obtain the bounding box of $y_{(j-1) \to j}$ as the location of instance $k$ in the $j$-th frame.
Similar to the Re-ID module, we extract the feature map according to this bounding box from $f_j$ by RoIAlign operation. The feature of this bounding box is denoted as $x_j$.
The historical information of instance $k$ from $i$-th frame to $(j-1)$-th frame is expressed by a hidden state or memory $h_{j-1} \in \mathbb{R}^{m \times m \times d}$, where $m\times m$ denotes the feature size and $d$ represents the number of channels.
We warp $h_{j-1}$ to $h_{(j-1) \to j}$ by optical flow for spatial consistency.
With both $x_j$ and $h_{(j-1) \to j}$ we can estimate ${h_j}$ by Eq.~\eqref{eqn:hidden_rnn}. Similar to the mask network described in Sec.~\ref{sec:re_identification}, we employ `conv5\_x' block of ResNet-101 as our recurrent function $\mathcal{N}_{R}$.
The mask for the instance $k$ in the $j$-th frame, $y_{j}$, can then be obtained by using the output function in Eq.~\eqref{eqn:output_rnn}. The output function $\mathcal{N}_{O}$ is modeled by three convolutional layers.

\begin{figure}[t]
    \centering
    \includegraphics[width=0.48\textwidth]{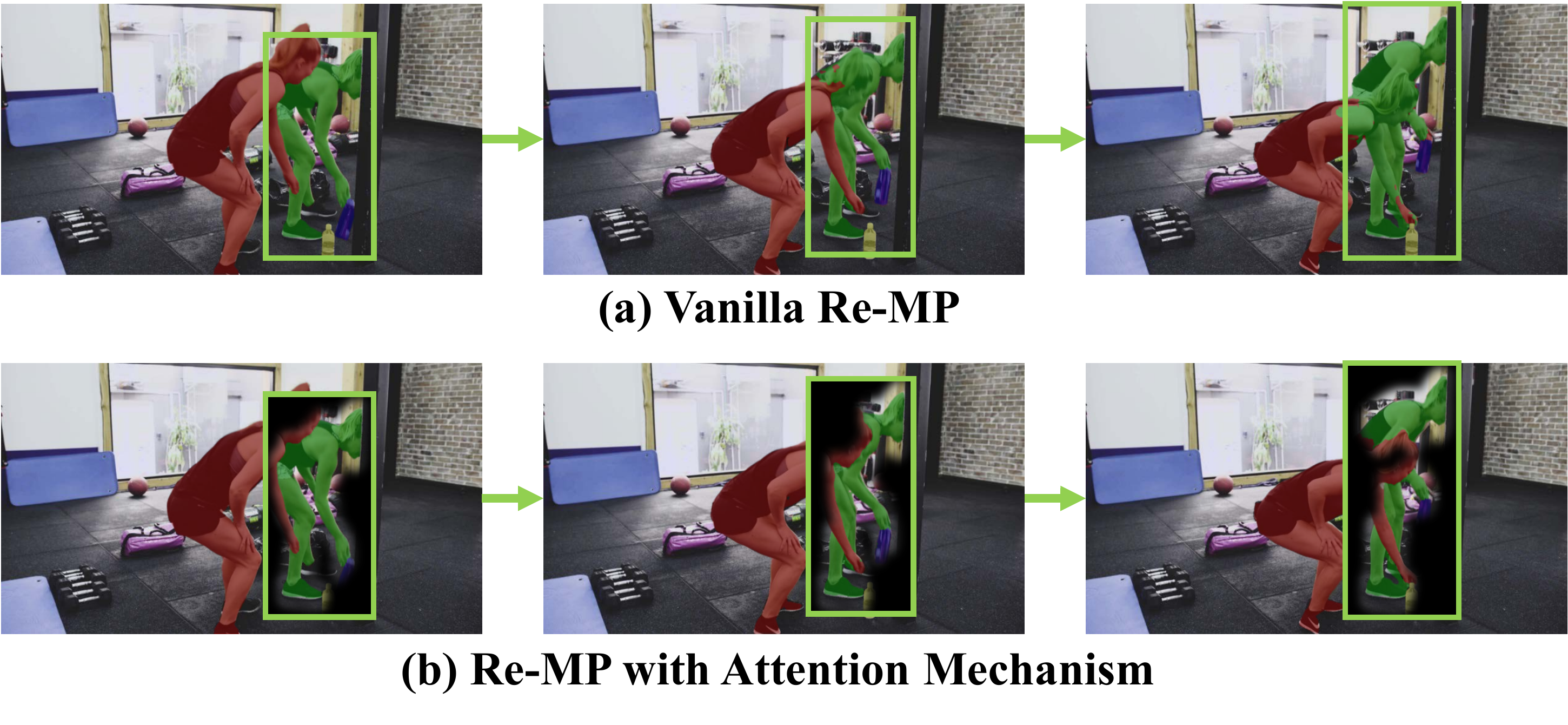}
    \vskip -0.3cm
    \caption{\small{Region attention in mask propagation.}}
    \label{fig:attention}
    \vspace{-16pt}
\end{figure}

\vspace{0.1cm}
\noindent
\textbf{Region attention.}
The quality of propagation to obtain $y_{j}$ relies on how accurate the model in capturing the shape of target instance. 
In many cases, a bounding box may contain distractors that can jeopardize the quality of mask propagated. 
As shown in Fig~\ref{fig:attention}(a), if we directly generate $y_{j}$ from $h_j$, a model is likely to be confused by distractors that appear in the bounding box.
To overcome this issue, we leverage the attention mechanism to filter out potentially noisy regions.
It is worth pointing out that attention mechanism has been used in various computer vision tasks~\cite{ba-attention-2015,yang2016stacked} but not mask propagation. Our work presents the first attempt to incorporate attention mechanism in mask propagation.

Specifically, given the warped hidden state $h_{(j-1) \to j}$ , we first feed it into a single convolutional layer and then a softmax function, to generate the attention distribution $a_{j} \in \mathbb{R}^{m \times m \times 1}$ over the bounding box.
Figure~\ref{fig:attention}(b) shows the attention distributions we learned.
Then we multiply the current hidden state $h_j$ by $a_{j}$ across all channels to focus on the regions we interested.
And the mask $y_{j}$ is generated from enhanced $h_j$ by using Eq.~\eqref{eqn:output_rnn}.
As shown in Fig.~\ref{fig:attention}, the Re-MP module concentrates on the tracked object thanks to the attention mechanism.
The mask propagation of an object aborts when its size is too small, indicating a high possibility of occlusion.
Finally, $\hat{y}$ is extended to a tracklet $\{y_{k1}, ...,y_{i+1},\hat{y},y_{i+1},..., y_{k2}\}$ after the forward and backward propagation.
%
%
This process is applied to all the starting points to generate a set of tracklets.
However, in some cases, different starting points may produce the same tracklet, which leads to redundant computation.
To speed up the algorithm, we sort all starting points descendingly by their cosine similarities against templates.
We extend the starting points according to the sorted order.
If a starting point's mask highly overlaps with a mask in existing tracklets, we skip this starting point.
This step does not affect the results; on the contrary, it greatly accelerates the inference speed.

\vspace{0.1cm}
\noindent
\textbf{Linking the tracklets}.
The previous mask propagation step generates potentially segmented tracklets.
We introduce a greedy approach to link those tracklets into consistent mask tubes.
It sorts all tracklets descendingly by cosine similarities between their respective starting point and templates.
Given the sorted order, tracklets with the highest similarities are assigned to the respective templates.
The method then examines the remaining tracklets in turn.
A tracklet is merged with a tracklet of higher order if there is no contradiction between them.
In practice, this simple mechanism works well. 
We will investigate other plausible linking approaches (\eg, conditional random field) in the future.

\subsection{Inference and Training}
\label{sec:infer_train}
\noindent
\textbf{Iterative inference.}
During inference, we are given a video sequence $\{I_1,...,I_N\}$, and the masks of target objects in the first frame.
As mentioned, we employ those masks as the initial templates.
DyeNet is iteratively applied to the whole video sequence until no more high confidence instances can be found.
The set of templates will be augmented by the predictions with high confidences after each iteration.

\noindent
\textbf{Training details.}
The overall loss function of DyeNet is formulated as:
$L = L_{reid} + \lambda (L_{mask} + L_{remp})$,
where $L_{reid}$ is the re-identification loss of re-identification network in Sec.~\ref{sec:re_identification}, which follows Online Instance Matching (OIM) loss in~\cite{xiao2017joint}.
$L_{mask}$ and $L_{remp}$ indicate the pixel-wise segmentation losses of the mask network in Sec.~\ref{sec:re_identification} and recurrent mask propagation module in Sec.~\ref{sec:recurrent_mp}.
%
%
The overall loss is a linear combination of those three losses, where $\mathbf{\lambda}$ is a weight that balances the scale of those lose terms.
Following~\cite{li2017video,khoreva2017lucid}, the weights of DyeNet are initialized by a semantic segmentation network~\cite{zhao2017pspnet}.
Due to memory limitation, the weights of `conv1' to `conv4\_20' are frozen during the training.
Following~\cite{li2017video}, we also pre-train the re-identification sub-network in Re-ID module on ImageNet~\cite{deng2009imagenet} dataset.
The DyeNet is than jointly trained on the DAVIS training sets using 24k iterations.
We fix a mini-batch size of $32$ images (from 8 videos, 4 frames for each video), momentum $0.9$ and weight decay of $5^{-4}$.
The initial learning rate is $10^{-3}$ and dropped by a factor of 10 after every 8k iterations.
For online training, we follow~\cite{khoreva2017lucid} to synthesize videos based on the first frame of testing videos and add them into the training set.

\section{Experiments}
\label{sec:experiments}

\noindent
\textbf{Datasets}.
To demonstrate the effectiveness and generalization ability of DyeNet, we evaluate our method on DAVIS 2016~\cite{perazzi2016benchmark}, DAVIS 2017~\cite{pont2017the}, SegTrack$_{\rm v2}$~\cite{FliICCV2013} and You-TubeObjects~\cite{PrestLCSF12} datasets.
DAVIS 2016 (DAVIS$_{\rm 16}$) dataset contains 50 high-quality video sequences (3455 frames) with all frames annotated with pixel-wise object masks.
Since DAVIS$_{\rm 16}$ focuses on single-object video segmentation, each video has only one foreground object.
There are 30 training and 20 validation videos.
DAVIS 2017 (DAVIS$_{\rm 17}$) supplements the training and validation sets of DAVIS$_{\rm 16}$ with 30 and 10 high-quality video sequences, respectively.
It also introduces another 30 development test videos and 30 challenge testing videos, which makes DAVIS$_{\rm 17}$ three times larger than its predecessor.
Besides that, DAVIS$_{\rm 17}$ re-annotates all video sequences with multiple objects.
All of these differences make it more challenging than DAVIS$_{\rm 16}$.
SegTrack$_{\rm v2}$ dataset contains 14 low resolution video sequences (947 frames) with 24 generic foreground objects.
For YouTubeObjects~\cite{PrestLCSF12} dataset, we consider a subset of 126 videos with around 20000 frames, and the pixel-level annotation are provided by~\cite{jain2014supervoxel}.

\noindent
\textbf{Evaluation metric}.
For DAVIS$_{\rm 17}$ dataset, we follow~\cite{pont2017the} that adopts region ($\mathcal{J}$), boundary ($\mathcal{F}$) and their average ($\mathcal{G}$) measures for evaluation.
To be consistent with existing studies~\cite{voigtlaender2017online,khoreva2017lucid,perazzi2017learning,caelles2017one}, we use mean intersection over union (mIoU) averaged across all instances to evaluate the performance in DAVIS$_{\rm 16}$, SegTrack$_{\rm v2}$ and YouTubeObjects.

\noindent
\textbf{Training modalities}.
In existing studies~\cite{perazzi2017learning,khoreva2017lucid}, training modalities can be divided into \textbf{offline training} and \textbf{online training}.
In offline training a model is only trained on the training set without any annotations from the test set.
Since the first frame annotations are provided in the testing stage, we can use them for tuning the model, namely online training.
Online training can be further divided into \textbf{per-dataset} and \textbf{per-video} training.
In per-dataset online training, we fine-tune a model based on all the first frame annotations from the test set, to obtain a dataset-specific model.
Per-video online training adapts the model weights to each testing video, \ie, there will be as many video specific models as the testing videos during the testing stage.

\subsection{Ablation Study}

In this section, we investigate the effectiveness of each component in DyeNet. Unless otherwise indicated we employ the \emph{train set} of DAVIS$_{\rm 17}$ for training. All performance are reported on the \emph{val set} of DAVIS$_{\rm 17}$. Offline training modality is used.

\noindent
\textbf{Effectiveness of Re-MP module}.
To demonstrate the effectiveness of Re-MP module clearly, we do not involve the Re-ID module in this experiment.
Re-MP module is directly applied to extend the annotations in the first frame to form mask tubes. This variant degenerates our method to a conventional mask propagation pipeline but with an attention-aware recurrent structure.
We compare Re-MP module with the state-of-the-art mask propagation method, MSK~\cite{perazzi2017learning}.
To ensure a fair comparison, we re-implement MSK to have the same backbone ResNet-101 as DyeNet.
We do not use online training and any post-processing in MSK either.
The re-implemented MSK achieves $78.7$ $\mathcal{J}$-mean on DAVIS$_{\rm 16}$ \emph{val set}, which is much higher than the original result $69.9$ reported in~\cite{perazzi2017learning}.
\begin{table}[h]
	\vspace{-8pt}
	\small
	\caption{Ablation study on Re-MP with DAVIS$_{\rm 17}$ \emph{val}.}
	\centering
	\begin{tabular}{@{}l@{\,}|c|@{}c@{\,}|@{}c@{\,}|@{}c@{\,}}
	\hline
		&Variant&~$\mathcal{J}$-mean&~$\mathcal{F}$-mean~&~$\mathcal{G}$-mean~\\
		\hline\hline
		MSK~\cite{perazzi2017learning}~&ResNet-101&63.3&67.2&65.3\\
		\hline
		\multirow{2}{*}{Re-MP~}	&no attention&65.3&69.7&67.5\\
								&full&\textbf{67.3}&\textbf{71.0}&\textbf{69.1}\\
	\hline
	\end{tabular}
	\label{tab:ablation_Re-MP}
\end{table}

\begin{figure}[b]
	\centering
	\includegraphics[width=0.48\textwidth]{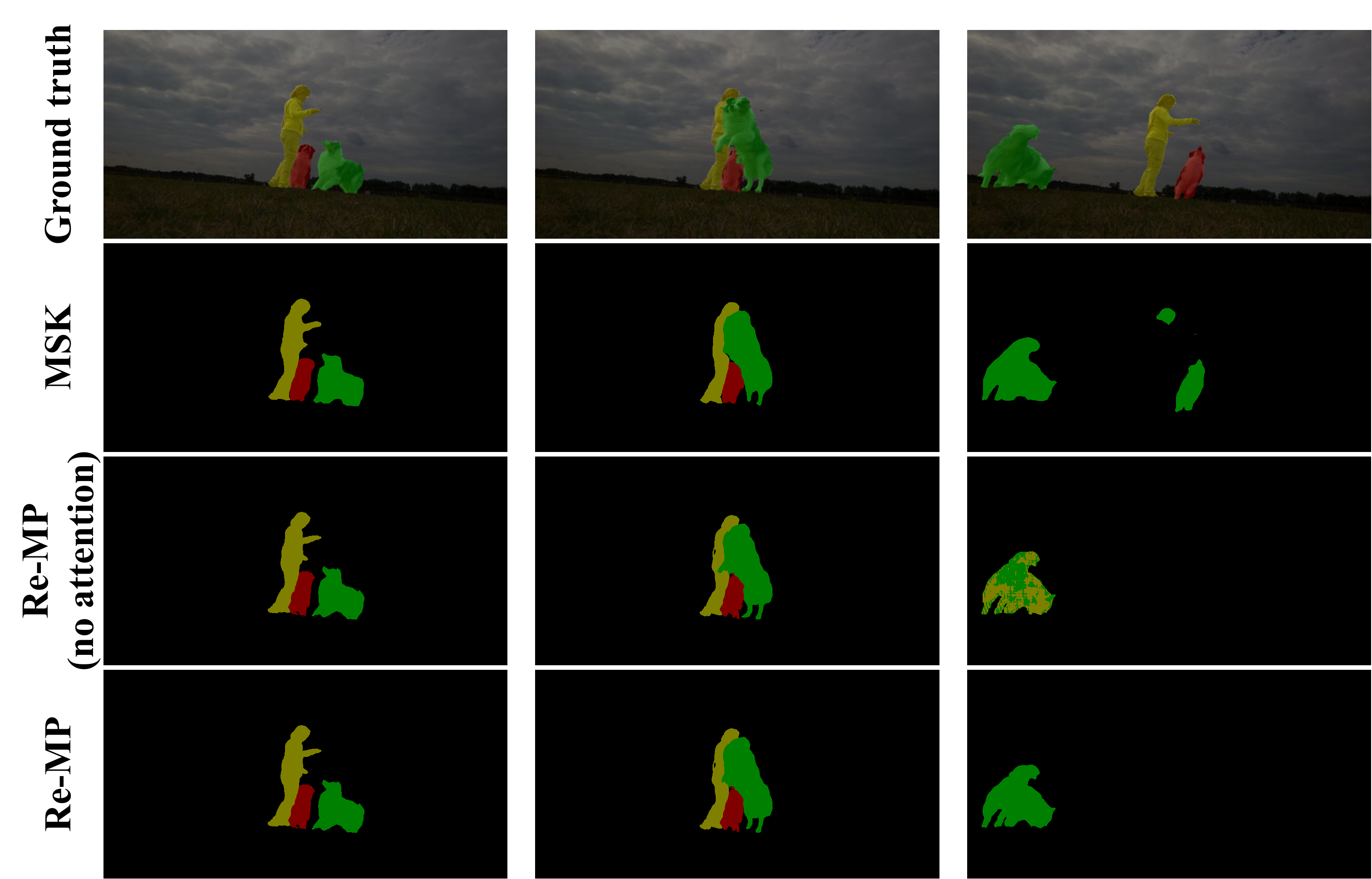}
	\caption{\small{Examples of mask propagation. \textbf{Best viewed in color.}}}
	\label{fig:ablation_Re-MP}
\end{figure}

\begin{table*}[t]
	\vspace{-8pt}
	\small
	\caption{Ablation study on Re-ID with DAVIS$_{\rm 17}$ \emph{val}. The improvement of $\mathcal{G}$-mean between rows is because of template expansion.}
	\centering
	\begin{tabular}{l|ccc|ccc|ccc|ccc}
		$\rho_{reid}$& \multicolumn{3}{c|}{0.9} & \multicolumn{3}{c|}{0.8} & \multicolumn{3}{c|}{0.7} & \multicolumn{3}{c}{0.6} \\
		\hline
		&preci.&recall&$\mathcal{G}$-mean&preci.&recall&$\mathcal{G}$-mean&preci.&recall&$\mathcal{G}$-mean&preci.&recall&$\mathcal{G}$-mean\\
		\hline
		\hline
		Iter. $1$&97.0&16.0&72.3&87.1&22.2&73.2&78.9&26.2&73.2&76.5&29.2&73.4\\
		Iter. $2$&90.3&29.3&73.3&75.6&32.5&73.7&68.9&33.5&\textbf{74.1}&65.5&34.1&74.0\\
		Iter. $3^+$&90.7&30.1&73.6&74.6&32.6&73.7&68.8&33.5&\textbf{74.1}&65.3&34.2&73.9
	\end{tabular}
	\label{tab:ablation_threshold}
\end{table*}

As shown in Table \ref{tab:ablation_Re-MP}, MSK achieves $65.3$ $\mathcal{G}$-mean on DAVIS$_{\rm 17}$ \emph{val set}.
Unlike MSK that propagates predicted masks only, the proposed Re-MP propagates all historical information by the recurrent architecture, and RoIAlign operation allows our network to focus on foreground regions and produce high-resolution masks, which makes Re-MP outperform MSK.
The Re-MP with attention mechanism is more focused on foreground regions, which further improves $\mathcal{G}$-mean by $1.6$.

Figure~\ref{fig:ablation_Re-MP} shows propagation results of different methods. In this video, a dog passes in front of a woman and another dog.
MSK dyes the woman and the back dog with the instance id of the front dog.
The plain Re-MP does not dye other instances, but it is still confused during the crossing and assigns the front dog with two instance ids.
Thanks to the attention mechanism, our full Re-MP is not distracted by other instances.
Due to occlusion, the masks of other instances are lost, and they will be retrieved by the Re-ID module in the complete DyeNet.

\noindent
\textbf{Effectiveness of Re-ID module with template expansion}.
In DyeNet, we employ the Re-ID module to search for target objects in the video sequence.
By choosing an appropriate similarity threshold $\rho_{reid}$, we can establish high-quality starting points for the Re-MP module.
The threshold $\rho_{reid}$ controls the trade-off between precision and recall of retrieved objects.
Table~\ref{tab:ablation_threshold} lists the precision and recall of retrieved starting points in each iteration as $\rho_{reid}$ varies, and corresponding overall performance. Tracklets are linked by greedy algorithm in this experiment.

Overall, the $\mathcal{G}$-mean is increased after each iteration due to the template expansion.
When $\rho_{reid}$ decreases, more instances are retrieved in the first iteration, which leads to high recall and $\mathcal{G}$-mean.
It also produces some imprecise starting points and further affects the quality of templates in subsequent iterations, so the increase of performance between each iteration is limited.
In contrast, Re-ID module with high $\rho_{reid}$ is stricter. As the template set expands, it can still achieve satisfying recall rate gradually.
In practice, the iterative process stops in about three rounds. Due to our greedy algorithm, the overall performance is less sensitive to $\rho_{reid}$. When $\rho_{reid} = 0.7$, DyeNet achieves the best $\mathcal{G}$-mean. This value is used in all the following experiments.

\noindent
\textbf{Effectiveness of each component in DyeNet}.
Table~\ref{tab:ablation} summarizes how performance gets improved by adding each component step-by-step into our DyeNet on the \emph{test-dev set} of DAVIS$_{\rm 17}$.
Our re-implemented MSK is chosen as the baseline.
All models in this experiment are first offline trained on the \emph{train} and \emph{val set}, and then per-dataset online trained on the \emph{test-dev set}.

\begin{table}
	\small
	\caption{Ablation study of each module in DyeNet with DAVIS$_{\rm 17}$ \emph{test-dev}.}
	\centering
	\begin{tabular}{@{}l@{\,}|c|@{}c@{\,}|@{}c@{\,}|@{}c@{\,}|@{}c@{\,}}
		\hline
		&Variant&~$\mathcal{J}$-mean&~$\mathcal{F}$-mean~&~$\mathcal{G}$-mean~&~$\Delta\mathcal{G}$-mean~\\
		\hline\hline
		MSK\cite{perazzi2017learning}~&ResNet-101&50.9&52.6&51.7&-\\
		\hline
		\multirow{2}{*}{Re-MP~}	&no attention&55.4&60.5&58.0&~+~6.2\\
								&full&59.1&62.8&61.0&~+~9.2\\
		\hline
		+~Re-ID~ &&\textbf{65.8}&\textbf{70.5}&\textbf{68.2}&~+~7.2\\
		\hline
		offline~&offline only&60.2&64.8&62.5&~-~5.6\\
		\hline
	\end{tabular}
	\vspace{-0.25cm}
	\label{tab:ablation}
\end{table}

Compared with MSK, our Re-MP module with attention mechanism significantly improves $\mathcal{G}$-mean by $9.2$.
The full DyeNet that contains both Re-ID and Re-MP modules achieves $68.2$ by using greedy algorithm to link the tracklets.
%
%
More remarkably, without online training, our DyeNet achieves a competitive $\mathcal{G}$-mean of $62.5$.

\begin{figure}
	\centering
	\includegraphics[width=0.48\textwidth]{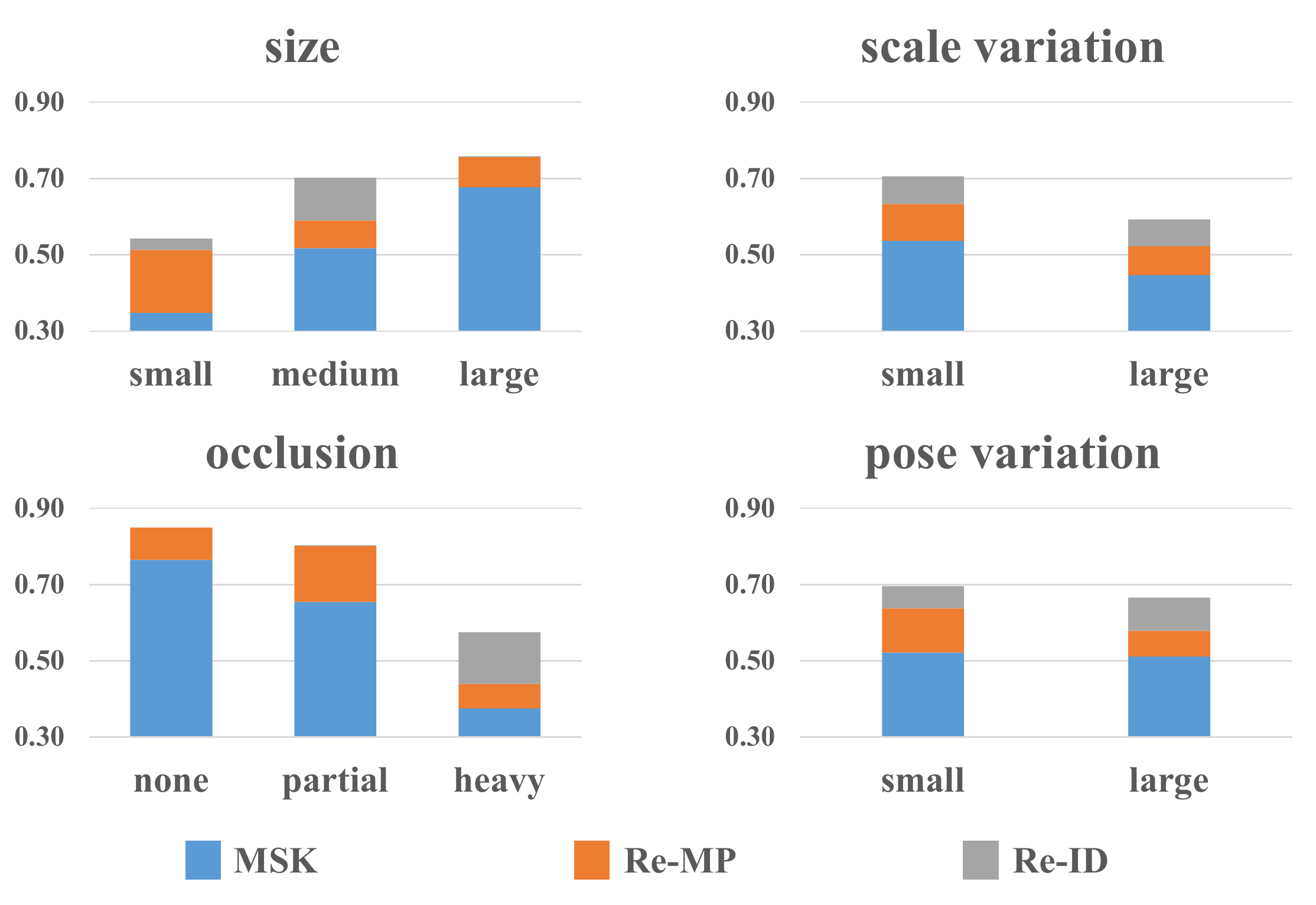}
	\vskip -0.25cm
	\caption{\small{Stage-wise performance increment according to specific attributes. \textbf{Best viewed in color.}}}
	\label{fig:attributes}
	\vspace{-10pt}
\end{figure}

To further investigate the contribution of each module in DyeNet, we categorize instances in \emph{test-dev set} by specific attributes, including:

\begin{packed_itemize}
	\small{
		\item \textbf{Size:}
		Instances are categorized into `small', `medium', and `large' according to their size in the first frames' annotations.
		\item \textbf{Scale Variation:}
		The area ratio among any pair of bounding boxes enclosing the target object is smaller than $0.5$. The bounding boxes are obtained from our best prediction.
		\item \textbf{Occlusion:}
		An object is not, partially, or heavily occluded.
		\item \textbf{Pose Variation:}
		Noticeable pose variation, due to object motion or relative camera-object rotation.
	}
\end{packed_itemize}

\begin{figure*}[t]
	\centering
	\includegraphics[width=1\textwidth]{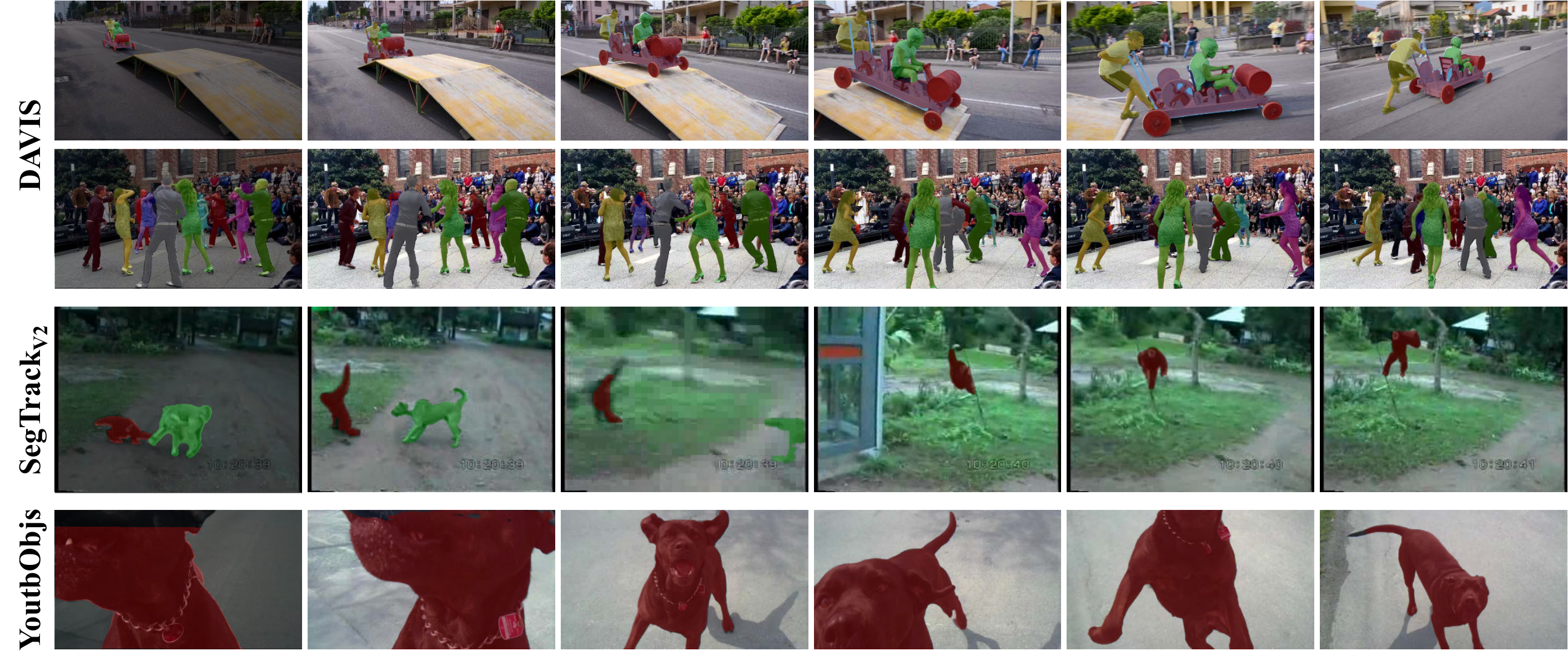}
	\vskip -0.3cm
	\caption{\small{Visualization of DyeNet's prediction. The first column shows the first frame of each video sequence with ground truth masks. The frames are chosen at equal interval. \textbf{Best viewed in color.}}}
	\label{fig:visualization}
\end{figure*}
We choose the best version of DyeNet in Table~\ref{tab:ablation}, and visualize its stage-wise performance according to specific attributes in Fig.~\ref{fig:attributes}.
We find that object's size and occlusion are most important factors that affect the performance, and scale variation has more influence on the performance than pose variation.
By inspecting closer, we observe that our Re-MP module can well track those small objects, which is the shortcoming of conventional mask propagation methods.
It also avoids our model being distracted by other objects in partial occlusion cases.
Complementary to Re-MP, Re-ID module retrieves missing instances due to heavy occlusions, greatly improves the performance in heavy occlusion cases.
Template expansion ensures Re-ID works well even if there are large pose variations.

\subsection{Benchmark}

In this section, we compare our DyeNet with other existing methods and show that it can achieve the state-of-the-art performance on standard benchmarks, including DAVIS$_{\rm 16}$, DAVIS$_{\rm 17}$, SegTrack$_{\rm v2}$ and YouTubeObjects datasets.
In this section, DyeNet is tested on a single scale without any post-processing.
Table~\ref{tab:davis_17} lists the $\mathcal{J}$, $\mathcal{F}$ and $\mathcal{G}$-means on DAVIS$_{\rm 17}$ \textit{test-dev}.
Approaches with ensemble are marked with $^\dagger$.
DyeNet is trained on \emph{train} and \emph{val sets} of DAVIS$_{\rm 17}$ and achieves a competitive $\mathcal{G}$-mean of $62.5$. It further improves $\mathcal{G}$-mean to $68.2$ through online fine-tuning, which is the best-performing method on DAVIS$_{\rm 17}$ benchmark.

\begin{table}[b]
	\small
	\caption{Results on DAVIS$_{\rm 17}$ \textit{test-dev}.}
	\centering
	\begin{tabular}{@{}l@{\,}|@{}c@{\,}|@{}c@{\,}|@{}c@{\,}|@{}c@{\,}|@{}c@{\,}}
		\hline
		\multirow{2}{*}{}&\multicolumn{2}{@{}c@{\,}|}{\footnotesize{~online training~}}&\multirow{2}{*}{\footnotesize{~$\mathcal{J}$-mean~}}&\multirow{2}{*}{\footnotesize{~$\mathcal{F}$-mean~}}&\multirow{2}{*}{\footnotesize{~$\mathcal{G}$-mean~}}\\ \cline{2-3}
		&\footnotesize{~dataset~}&\footnotesize{~video~}&&&\\
		\hline\hline
		OnAVOS$^\dagger$\cite{voigtlaender2017online}~&${\surd}$&${\surd}$&53.4&59.6&56.5\\
		LucidTracker\cite{khoreva2017lucid}~&${\surd}$&${\surd}$&60.1&68.3&64.2\\
		VS-ReID\cite{li2017video}~&${\surd}$&${\times}$&64.4&67.8&66.1\\
		LucidTracker$^\dagger$\cite{khoreva2017lucid}~&${\surd}$&${\surd}$&63.4&69.9&66.6\\
		\hline
		DyeNet~(offline)~&${\times}$&${\times}$&60.2&64.8&62.5\\
		DyeNet&${\surd}$&${\times}$&\textbf{65.8}&\textbf{70.5}&\textbf{68.2}\\
		\hline
	\end{tabular}
	\label{tab:davis_17}
\end{table}
\begin{table}[t]
	\small
	\caption{Results (mIoU) across three datasets.}
	\centering
	\begin{tabular}{@{}l@{\,}|@{}c@{\,}|@{}c@{\,}|@{}c@{\,}|@{}c@{\,}|@{}c@{\,}}
		\hline
		\multirow{2}{*}{}&\multicolumn{2}{@{}c@{\,}|}{\footnotesize{~online training~}}&\multirow{2}{*}{\footnotesize{~DAVIS$_{\rm 16}$~}}&\multirow{2}{*}{\footnotesize{~SegTrack$_{\rm v2}$~}}&\multirow{2}{*}{\footnotesize{~YoutbObjs~}}\\ \cline{2-3}
		&\footnotesize{~dataset~}&\footnotesize{~video~}&&&\\
		\hline\hline
		VPN\cite{jampani2017video}~&${\times}$&${\times}$&75.0&-&-\\
		SegFlow\cite{Cheng_ICCV_2017}~&${\surd}$&${\surd}$&76.1&-&-\\
		OSVOS\cite{caelles2017one}~&${\surd}$&${\surd}$&79.8&65.4&72.5\\
		MSK\cite{perazzi2017learning}~&${\surd}$&${\surd}$&80.3&70.3&72.6\\
		LucidTracker\cite{khoreva2017lucid}~&${\surd}$&${\surd}$&84.8&77.6&76.2\\
		OnAVOS\cite{voigtlaender2017online}~&${\surd}$&${\surd}$&85.7&-&77.4\\
		\hline
		DyeNet~(offline)~&${\times}$&${\times}$&84.7&\textbf{78.3}&74.9\\
		DyeNet&${\surd}$&${\times}$&\textbf{86.2}&\textbf{78.7}&\textbf{79.6}\\
		\hline
	\end{tabular}
	\vspace{-5pt}
	\label{tab:other_datasets}
\end{table}

To show the generalization ability and transferability of DyeNet, we next evaluate DyeNet on three other benchmarks, DAVIS$_{\rm 16}$, SegTrack$_{\rm v2}$ and YouTubeObjects, which contain diverse videos.
For DAVIS$_{\rm 16}$, DyeNet is trained on its \emph{train set}.
Since there is no video for offline training in SegTrack$_{\rm v2}$ and YouTubeObjects, we directly employ the model of DAVIS$_{\rm 17}$ as their offline model.
As summarized in Table \ref{tab:other_datasets}, offline DyeNet obtains promising performance, and after online fine-tuning, our model achieves state-of-the-art performance on all three datasets.
Note that although the videos in SegTrack$_{\rm v2}$ and YouTubeObjects are very different from videos in DAVIS$_{\rm 17}$, DyeNet trained on DAVIS$_{\rm 17}$ still gains outstanding performance on those datasets without any fine-tuning, which shows its great generalization ability and transferability to diverse videos.
We also find that our offline predictions on YouTubeObjects are even better than most ground-truth annotations, and performance losses are mainly caused by annotation bias.
In Fig.~\ref{fig:visualization}, we demonstrate some examples of DyeNet's predictions.

\noindent
\textbf{Speed Analysis}.
Most of existing methods require online training with post-processing to achieve a competitive performance. Because of those time consuming processes, their speed of inference is slow.
For example, the full OnAVOS~\cite{voigtlaender2017online} takes roughly 13 seconds per frame to achieve $85.7$ mIoU on DAVIS$_{\rm 16}$ \emph{val set}.
LucidTracker~\cite{khoreva2017lucid} that achieves $84.8$ mIoU requires 40k iterations per-dataset, 2k per-video online training and post-processing\cite{felzenszwalb2006efficient}.
Our offline DyeNet is capable of obtaining similar performance ($84.7$ mIoU) at 2.4 FPS on a single Titan Xp GPU.
After 2k per-dataset online training, our DyeNet achieves $86.2$ mIoU, and the corresponding running time is 0.43 FPS.

\section{Conclusion}
We have presented DyeNet, which joints re-identification and attention-based recurrent temporal propagation into a unified framework to address challenging video object segmentation with multiple instances.
This is the first end-to-end framework for this problem with a few compelling components.
First, to cope with pose variations of targets, we relaxed the reliance of template set in the first frame by performing template expansion in our iterative algorithm. Second, to achieve robust video segmentation against distractors and background clutters, we proposed attention mechanism for recurrent temporal propagation.
DyeNet does not require online training to obtain competitive accuracies at a faster speed than many existing methods.
With online training, DyeNet achieves state-of-the-art performance on a wide range of standard benchmarks (including DAVIS, SegTrack$_{\rm v2}$ and YouTubeObjects).

{\small
\bibliographystyle{ieee}
\bibliography{short,egbib}

\begin{thebibliography}{10}\itemsep=-1pt

\bibitem{ba-attention-2015}
J.~Ba, V.~Mnih, and K.~Kavukcuoglu.
\newblock Multiple object recognition with visual attention.
\newblock In {\em ICLR}. 2015.

\bibitem{bolme2010visual}
D.~S. Bolme, J.~R. Beveridge, B.~A. Draper, and Y.~M. Lui.
\newblock Visual object tracking using adaptive correlation filters.
\newblock In {\em CVPR}, 2010.

\bibitem{caelles2017one}
S.~Caelles, K.-K. Maninis, J.~Pont-Tuset, L.~Leal-Taix\'e, D.~Cremers, and
  L.~Van~Gool.
\newblock One-shot video object segmentation.
\newblock In {\em CVPR}, 2017.

\bibitem{chen2014semantic}
L.-C. Chen, G.~Papandreou, I.~Kokkinos, K.~Murphy, and A.~L. Yuille.
\newblock Semantic image segmentation with deep convolutional nets and fully
  connected crfs.
\newblock In {\em ICLR}, 2015.

\bibitem{Cheng_ICCV_2017}
J.~Cheng, Y.-H. Tsai, S.~Wang, and M.-H. Yang.
\newblock Segflow: Joint learning for video object segmentation and optical
  flow.
\newblock In {\em ICCV}, 2017.

\bibitem{deng2009imagenet}
J.~Deng, W.~Dong, R.~Socher, L.-J. Li, K.~Li, and L.~Fei-Fei.
\newblock Imagenet: A large-scale hierarchical image database.
\newblock In {\em CVPR}, 2009.

\bibitem{felzenszwalb2006efficient}
P.~F. Felzenszwalb and D.~P. Huttenlocher.
\newblock Efficient belief propagation for early vision.
\newblock {\em IJCV}, 70(1):41--54, 2006.

\bibitem{grundmann2010efficient}
M.~Grundmann, V.~Kwatra, M.~Han, and I.~Essa.
\newblock Efficient hierarchical graph-based video segmentation.
\newblock In {\em CVPR}, 2010.

\bibitem{hariharan2014simultaneous}
B.~Hariharan, P.~Arbel{\'a}ez, R.~Girshick, and J.~Malik.
\newblock Simultaneous detection and segmentation.
\newblock In {\em ECCV}, 2014.

\bibitem{hariharan2015hypercolumns}
B.~Hariharan, P.~Arbel{\'a}ez, R.~Girshick, and J.~Malik.
\newblock Hypercolumns for object segmentation and fine-grained localization.
\newblock In {\em CVPR}, 2015.

\bibitem{he2017mask}
K.~He, G.~Gkioxari, P.~Doll{\'a}r, and R.~Girshick.
\newblock Mask {R-CNN}.
\newblock In {\em ICCV}, 2017.

\bibitem{he2016deep}
K.~He, X.~Zhang, S.~Ren, and J.~Sun.
\newblock Deep residual learning for image recognition.
\newblock In {\em CVPR}, 2016.

\bibitem{ilg2017flownet}
E.~Ilg, N.~Mayer, T.~Saikia, M.~Keuper, A.~Dosovitskiy, and T.~Brox.
\newblock {FlowNet} 2.0: Evolution of optical flow estimation with deep
  networks.
\newblock In {\em CVPR}, 2017.

\bibitem{jain2014supervoxel}
S.~D. Jain and K.~Grauman.
\newblock Supervoxel-consistent foreground propagation in video.
\newblock In {\em ECCV}, 2014.

\bibitem{jampani2017video}
V.~Jampani, R.~Gadde, and P.~V. Gehler.
\newblock Video propagation networks.
\newblock In {\em CVPR}, 2017.

\bibitem{khoreva2017lucid}
A.~Khoreva, R.~Benenson, E.~Ilg, T.~Brox, and B.~Schiele.
\newblock Lucid data dreaming for object tracking.
\newblock In {\em CVPRW}, 2017.

\bibitem{nguyen2017instance}
T.-N. Le, K.-T. Nguyen, M.-H. Nguyen-Phan, T.-V. Ton, T.-A. Nguyen, X.-S.
  Trinh, Q.-H. Dinh, V.-T. Nguyen, A.-D. Duong, A.~Sugimoto, T.~V. Nguyen, and
  M.-T. Tran.
\newblock Instance re-identification flow for video object segmentation.
\newblock In {\em CVPRW}, 2017.

\bibitem{lee2011key}
Y.~J. Lee, J.~Kim, and K.~Grauman.
\newblock Key-segments for video object segmentation.
\newblock In {\em ICCV}, 2011.

\bibitem{FliICCV2013}
F.~Li, T.~Kim, A.~Humayun, D.~Tsai, and J.~M. Rehg.
\newblock Video segmentation by tracking many figure-ground segments.
\newblock In {\em ICCV}, 2013.

\bibitem{li2017not}
X.~Li, Z.~Liu, P.~Luo, C.~C. Loy, and X.~Tang.
\newblock Not all pixels are equal: difficulty-aware semantic segmentation via
  deep layer cascade.
\newblock In {\em CVPR}, 2017.

\bibitem{li2017video}
X.~Li, Y.~Qi, Z.~Wang, K.~Chen, Z.~Liu, J.~Shi, P.~Luo, X.~Tang, and C.~C. Loy.
\newblock Video object segmentation with re-identification.
\newblock In {\em CVPRW}, 2017.

\bibitem{li2016fully}
Y.~Li, H.~Qi, J.~Dai, X.~Ji, and Y.~Wei.
\newblock Fully convolutional instance-aware semantic segmentation.
\newblock In {\em CVPR}, 2017.

\bibitem{liu2017deep}
Z.~Liu, X.~Li, P.~Luo, C.~C. Loy, and X.~Tang.
\newblock Deep learning markov random field for semantic segmentation.
\newblock {\em TPAMI}, 2017.

\bibitem{marki2016bilateral}
N.~M{\"a}rki, F.~Perazzi, O.~Wang, and A.~Sorkine-Hornung.
\newblock Bilateral space video segmentation.
\newblock In {\em CVPR}, 2016.

\bibitem{papazoglou2013fast}
A.~Papazoglou and V.~Ferrari.
\newblock Fast object segmentation in unconstrained video.
\newblock In {\em ICCV}, 2013.

\bibitem{perazzi2017learning}
F.~Perazzi, A.~Khoreva, R.~Benenson, B.~Schiele, and A.~Sorkine-Hornung.
\newblock Learning video object segmentation from static images.
\newblock In {\em CVPR}, 2017.

\bibitem{perazzi2016benchmark}
F.~Perazzi, J.~Pont-Tuset, B.~McWilliams, L.~Van~Gool, M.~Gross, and
  A.~Sorkine-Hornung.
\newblock A benchmark dataset and evaluation methodology for video object
  segmentation.
\newblock In {\em CVPR}, 2016.

\bibitem{pont2017the}
J.~Pont-Tuset, F.~Perazzi, S.~Caelles, P.~Arbel\'aez, A.~Sorkine-Hornung, and
  L.~Van~Gool.
\newblock The 2017 davis challenge on video object segmentation.
\newblock {\em arXiv:1704.00675}, 2017.

\bibitem{PrestLCSF12}
A.~Prest, C.~Leistner, J.~Civera, C.~Schmid, and V.~Ferrari.
\newblock Learning object class detectors from weakly annotated video.
\newblock In {\em CVPR}, 2012.

\bibitem{ren2015faster}
S.~Ren, K.~He, R.~Girshick, and J.~Sun.
\newblock Faster {R-CNN}: Towards real-time object detection with region
  proposal networks.
\newblock In {\em NIPS}, 2015.

\bibitem{tsai2016video}
Y.-H. Tsai, M.-H. Yang, and M.~J. Black.
\newblock Video segmentation via object flow.
\newblock In {\em CVPR}, 2016.

\bibitem{valmadre2017end}
J.~Valmadre, L.~Bertinetto, J.~F. Henriques, A.~Vedaldi, and P.~H. Torr.
\newblock End-to-end representation learning for correlation filter based
  tracking.
\newblock 2017.

\bibitem{voigtlaender2017online}
P.~Voigtlaender and B.~Leibe.
\newblock Online adaptation of convolutional neural networks for video object
  segmentation.
\newblock In {\em BMVC}, 2017.

\bibitem{xiao2016track}
F.~Xiao and Y.~Jae~Lee.
\newblock Track and segment: An iterative unsupervised approach for video
  object proposals.
\newblock In {\em CVPR}, 2016.

\bibitem{xiao2017joint}
T.~Xiao, S.~Li, B.~Wang, L.~Lin, and X.~Wang.
\newblock Joint detection and identification feature learning for person
  search.
\newblock In {\em CVPR}, 2017.

\bibitem{xu2012streaming}
C.~Xu, C.~Xiong, and J.~J. Corso.
\newblock Streaming hierarchical video segmentation.
\newblock In {\em ECCV}, 2012.

\bibitem{yang2016stacked}
Z.~Yang, X.~He, J.~Gao, L.~Deng, and A.~Smola.
\newblock Stacked attention networks for image question answering.
\newblock In {\em CVPR}, 2016.

\bibitem{yoon2017pixel}
J.~S. Yoon, F.~Rameau, J.~Kim, S.~Lee, S.~Shin, and I.~S. Kweon.
\newblock Pixel-level matching for video object segmentation using
  convolutional neural networks.
\newblock In {\em CVPR}, 2017.

\bibitem{zhao2017pspnet}
H.~Zhao, J.~Shi, X.~Qi, X.~Wang, and J.~Jia.
\newblock Pyramid scene parsing network.
\newblock In {\em CVPR}, 2017.

\bibitem{crfasrnn_iccv2015}
S.~Zheng, S.~Jayasumana, B.~Romera-Paredes, V.~Vineet, Z.~Su, D.~Du, H.~Chang,
  and P.~Torr.
\newblock Conditional random fields as recurrent neural networks.
\newblock In {\em ICCV}, 2015.

\end{thebibliography}
}

\end{document}